\documentclass[10pt,twocolumn,letterpaper]{article}

\usepackage{iccv}
\usepackage{times}
\usepackage{epsfig}
\usepackage{graphicx}
\usepackage{amsmath}
\usepackage{amssymb}

\usepackage{graphicx}
\usepackage{wrapfig}
\usepackage{amsmath}
\usepackage{amssymb}
\usepackage{booktabs}
\usepackage{threeparttable}
\usepackage{float}
\usepackage{verbatim}
\usepackage[dvipsnames]{xcolor}
\usepackage{comment}
\newcommand{\ve}[1]{\mathbf{#1}} 
\usepackage{amsmath}  
\usepackage{algorithm}
\usepackage{algorithmic}
\usepackage{colortbl}
\usepackage{diagbox}
\usepackage[accsupp]{axessibility}  
\definecolor{darkseagreen}{rgb}{0.56, 0.74, 0.56}
\definecolor{lightpink}{rgb}{1.0, 0.71, 0.76}
\usepackage{bbm}
\usepackage{paralist,bbding,pifont}

\usepackage{multirow}

\definecolor{citecolor}{HTML}{0071BC}
\definecolor{linkcolor}{HTML}{ED1C24}
\usepackage[pagebackref=false, breaklinks=true, letterpaper=true, colorlinks, citecolor=citecolor, linkcolor=linkcolor, bookmarks=false]{hyperref}

\usepackage[capitalize]{cleveref}
\crefname{section}{Sec.}{Secs.}
\Crefname{section}{Section}{Sections}
\Crefname{table}{Table}{Tables}
\crefname{table}{Tab.}{Tabs.}
\def\eg{\emph{e.g}\onedot} 
\def\ie{\emph{i.e}\onedot} 
 
\def\etc{\emph{etc}\onedot}

\makeatother

\iccvfinalcopy 

\def\iccvPaperID{****} 

\begin{document}
\title{Open-vocabulary Panoptic Segmentation with Embedding Modulation}
\author{Xi Chen$^{1}$\quad Shuang Li$^{2}$\quad Ser-Nam Lim$^{3}$\quad Antonio Torralba$^{2}$\quad Hengshuang Zhao$^{1,2}$\\
$^{1}$The University of Hong Kong \quad $^{2}$Massachusetts Institute of Technology \quad $^{3}$Meta AI\\}
\maketitle
\begin{abstract}
Open-vocabulary image segmentation is attracting increasing attention due to its critical applications in the real world. Traditional closed-vocabulary segmentation methods are not able to characterize novel objects, whereas several recent open-vocabulary attempts obtain unsatisfactory results, \ie, notable performance reduction on the closed-vocabulary and massive demand for extra data. To this end, we propose \textbf{OPSNet}, an omnipotent and data-efficient framework for \textbf{O}pen-vocabulary \textbf{P}anoptic \textbf{S}egmentation. 
Specifically, the exquisitely designed Embedding Modulation module, together with several meticulous components, enables adequate embedding enhancement and information exchange between the segmentation model and the visual-linguistic well-aligned CLIP encoder, resulting in superior segmentation performance under both open- and closed-vocabulary settings with much fewer need of additional data. Extensive experimental evaluations are conducted across multiple datasets (\eg, COCO, ADE20K, Cityscapes, and PascalContext) under various circumstances, where the proposed OPSNet achieves state-of-the-art results, which demonstrates the effectiveness and generality of the proposed approach. The code and trained models will be made publicly available.
\end{abstract}

\section{Introduction}
\vspace{-3pt}
The real world is diverse and contains numerous distinct objects. In practical scenarios, we inevitably encounter various objects with different shapes, colors, and categories. Although some of them are unfamiliar or rarely seen, to better understand the world, we still need to figure out the region and shape of each object and what it is. The ability to perceive and segment both known and unknown objects is natural and essential for many real-world applications like autonomous driving, robot sensing and navigation, human-object interaction, augmented reality, healthcare, \etc.

\begin{figure}[t]
\centering 
\includegraphics[width=1.0\linewidth]{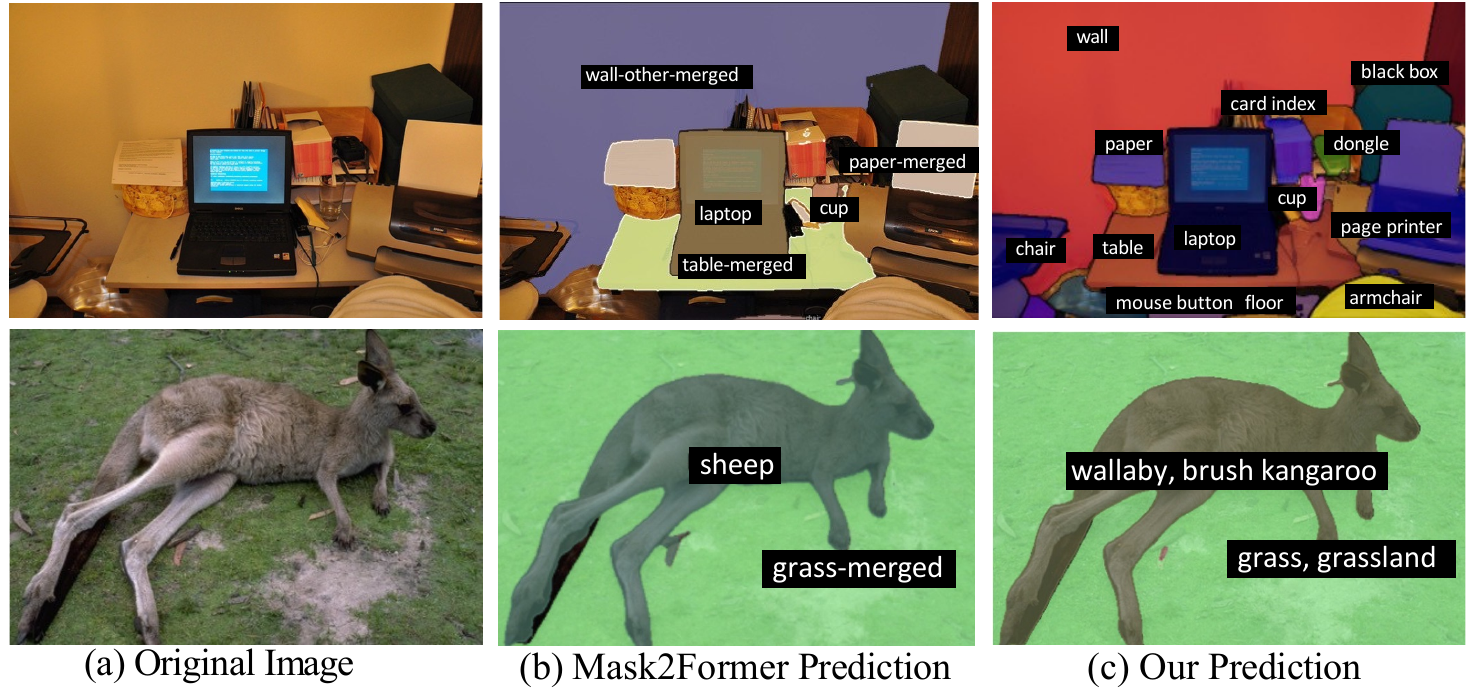} 
\vspace{-10pt}
\caption{Visual comparisons of classical closed-vocabulary segmentation and our open-vocabulary segmentation. Models are trained on the COCO panoptic dataset. Categories like `printer', `card index', `dongle', and `kangaroo' are not presented in the COCO concept set. Closed-vocabulary segmentation algorithms like Mask2Former~\cite{mask2former} are not able to detect and segment new objects (top middle) or fail to recognize object categories (bottom middle). In contrast, our approach is able to segment and recognize novel objects (top right, bottom right) for the open vocabulary.}
\vspace{-10pt}
\label{fig:1}
\end{figure}

Lots of works have explored image segmentation and achieved great success~\cite{pspnet,maskrcnn,knet,mask2former}. However, they are typically designed and developed on specific datasets (\eg, COCO~\cite{coco}, ADE20K~\cite{ade20k}) with predefined categories in a closed vocabulary, which assume the data distribution and category space remain unchanged during algorithm development and deployment procedures, resulting in noticeable and unsatisfactory failures when handling new environments in the complex real world, as shown in Fig.~\ref{fig:1} (b).

To address this problem, open-vocabulary perception is densely explored for semantic segmentation and object detection. Some methods~\cite{openseg,Detic,ViLD,GLIP,GLIPv2} use the visual-linguistic well-aligned CLIP~\cite{clip} text encoder to extract the language embeddings of category names to represent each category, and train the classification head to match these language embeddings. However, training the text-image alignment from scratch often requires a large amount of data and a heavy training burden. 
Other works~\cite{MaskCLIP,simplebaselinezsg} use both of the pre-trained CLIP image/text encoders to transfer the open-vocabulary ability from CLIP. However, as CLIP is not a cure-all for all domains and categories, although they are data-efficient, they struggle to balance the generalization ability and the performance in the training domain. \cite{simplebaselinezsg, zegformer} demonstrate suboptimal cross-datasset results, \cite{MaskCLIP} shows unsatifactory performance on the training domain. 
Besides, their methods for leveraging CLIP visual features are inefficient. Specifically, they need to pass each of the proposals into the CLIP image encoder to extract the visual features. 

Considering the characteristics and challenges of the previous methods, we propose \textbf{OPSNet} for \textit{\textbf{O}pen-vocabulary \textbf{P}anoptic \textbf{S}egmentation}, which is omnipotent and data-efficient for both open- and closed-vocabulary settings. 
Given an image, OPSNet first predicts class-agnostic masks for all objects and learns a series of in-domain query embeddings. 
For classification, a Spatial Adapter is added after the CLIP image encoder to maintain the spatial resolution. Then Mask Pooling uses the class-agnostic masks to pool the visual feature into CLIP embeddings, thus the visual embedding for each object can be extracted in one pass.

Afterward, we propose the key module named Embedding Modulation to produce the modulated embeddings for classification according to the query embeddings, CLIP embeddings, and the concept semantics.
This modulated final embedding could be used to match the text embeddings of category names extracted by the CLIP text encoder. 
Embedding Modulation combines the advantages of query and CLIP embeddings, and enables adequate embedding enhancement and information exchange for them, thus making OPSNet omnipotent for generalized domains and data-efficient for training.
To further push the boundary of our framework, we propose Mask Filtering to improve the quality of mask proposals, and Decoupled Supervision to scale up the training concepts using image-level labels to train classification and the self-constrains to supervise masks.

With these exquisite designs, OPSNet archives superior performance on COCO~\cite{coco}, shows exceptional cross-dataset performance on ADE20K~\cite{ade20k}, Cityscapes~\cite{cityscape}, PascalContext~\cite{pascalcontext}, and generalizes well to novel objects in the open vocabulary, as shown in Fig.~\ref{fig:1} (c).

In general, our contributions could be summarized as:

\begin{compactitem}
    \item We address the challenging open-vocabulary panoptic segmentation task and propose a novel framework named OPSNet, which is omnipotent and data-efficient, with the assistance of the exquisitely designed Embedding Modulation module.
    \item We propose several meticulous components like Spatial Adapter, Mask Pooling, Mask Filtering, and Decoupled Supervision, which are proven to be of great benefit for open-vocabulary segmentation.
    \item We conduct extensive experimental evaluations across multiple datasets under various circumstances, and the harvested state-of-the-art results demonstrate the effectiveness and generality of the proposed approach.
\end{compactitem}
\vspace{-5pt}
\section{Related Work}
\vspace{-5pt}
\paragraph{Unified image segmentation.}
Image segmentation targets grouping coherent pixels. Classical model architectures for semantic~\cite{fcn,deeplabv2,pspnet,psanet,ocrnet}, instance~\cite{maskrcnn,panet,hybridtaskcascade,yolact,tian2020conditional}, and panoptic~\cite{panopticseg,upsnet,Panoptic-deeplab,panopticfpn,panopticfcn} segmentation differ greatly. Recently, some works~\cite{maxdeeplab,knet,maskformer,mask2former} propose unified frameworks for image segmentation. With the help of vision transformers~\cite{vit,liu2021swin,DETR}, they retain a set of learnable queries, use these queries as convolutional kernels to produce multiple binary masks, and add an MLP head on the updated queries to predict the categories of the binary masks. This kind of simple pipeline is suitable for different segmentation tasks, and is called unified image segmentation. Nevertheless, although they design a universal structure, they are developed on specific datasets with predefined categories. Once trained on a dataset, these models could only conduct segmentation within the predefined categories in a closed vocabulary, resulting in inevitable failures in the real open vocabulary. 
We extend their scope to open vocabulary. Our model provides not only an omnipotent structure for different segmentation tasks, but also an omnipotent recognition ability for diverse scenarios in open vocabulary.

\vspace{-12pt}
\paragraph{Class-agnostic detection and segmentation.}
To generalize the localization ability of the existing detection and segmentation models, some works~\cite{entityseg,OLN,withoutclassify,uvo,konan2022extending} remove the classification head of a detection or segmentation model and treat all categories as entities. It is proven that the class-agnostic models can detect more objects since they focus on learning the generalizable knowledge of `what makes an object' rather than distinguishing visually similar classes like `house' or `building', and `cow' or `sheep', etc.  Although they give better mask predictions for general categories, recognizing the detected objects is not touched. 

\vspace{-12pt}
\paragraph{Open-vocabulary detection and segmentation.}

Some recent works try to tackle open-vocabulary detection and segmentation using language embeddings. \cite{GLIP,GLIPv2} leverage the large-scale image-text pairs to pre-train the detection network. ViLD~\cite{ViLD} distills the knowledge of ALIGN~\cite{ALIGN} to improve the detector's generalization ability. Detic~\cite{Detic} utilizes the ImageNet-21K\cite{imagenet} data to expand the detection categories. For segmentation, \cite{simplebaselinezsg} proposes a two-stage pipeline, where generalizable mask proposals are extracted and then fed into CLIP~\cite{clip} for classification. DenseCLIP~\cite{denseclipccl} adopts text embedding as a classifier to conduct convolution on feature maps produced by CLIP image encoder, and extends the architecture of the image encoder to semantic segmentation models~\cite{pspnet,deeplab}. OpenSeg~\cite{openseg} predicts general mask proposals and aligns the mask pooled features to the language space of ALIGN~\cite{ALIGN} with large-scale caption data~\cite{localizednarratives} for training. They reach great zero-shot performance for a large range of categories. However, all these works~\cite{denseclipccl,openseg,simplebaselinezsg} only deal with semantic segmentation. OpenSeg~\cite{openseg} and \cite{simplebaselinezsg} predict general masks that are noisy and overlapped, which could not accomplish instance-level distinction. MaskCLIP~\cite{MaskCLIP} is the only existing work for panoptic segmentation, which trains to gather the feature from a pre-trained CLIP image encoder. However, although it reaches great cross-dataset ability, its performance on COCO is far from satisfactory.  

\begin{figure*}[t]
\centering 
\includegraphics[width=1.0\linewidth]{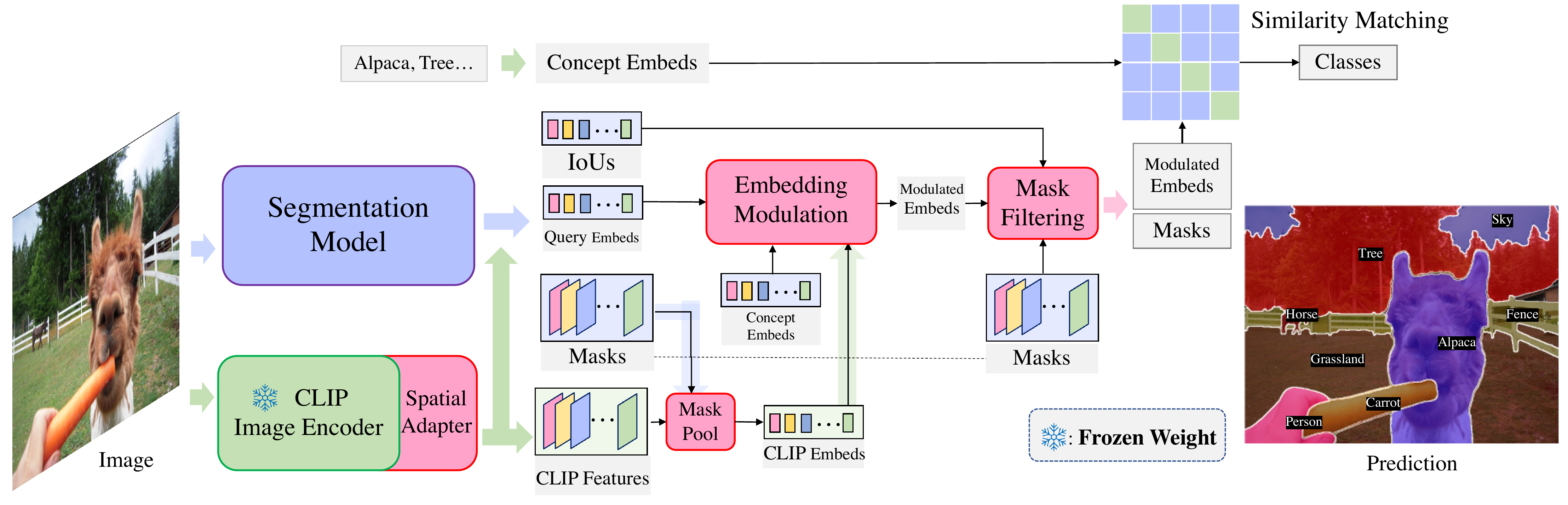} 
\caption{The overall pipeline of OPSNet, our novel blocks are marked in red. For an input image, the segmentation model predicts updated query embeddings, binary masks, and IoU scores. Meanwhile, we leverage a Spatial Adapter to extract CLIP visual features. We use these CLIP features to enhance the query embeddings and use binary masks to pool them into CLIP embeddings. Afterward, the CLIP Embed, Query Embeds, and Concept Embeds are fed into the Embedding Modulation module to produce the  modulated embeddings.  Next, we use Mask Filtering to remove low-quality proposals thus getting masks and embeddings for each object. Finally, 
we use the modulated embeddings to match the text embeddings extracted by the CLIP text encoder and assign a category label for each mask.
}
\label{fig:pipeline}
\vspace{-10pt}
\end{figure*}

\section{Method}

We introduce our OPSNet, an omnipotent and data-efficient framework for open-vocabulary panoptic segmentation. The overall pipeline is demonstrated in Fig.~\ref{fig:pipeline}. 
 We introduce our roadmap towards open-vocabulary  from the vanilla version to our exquisite designs.

\subsection{Vanilla Open-vocabulary Segmentation}
\label{sec:baseline}

Inspired by DETR~\cite{DETR}, recently, unified image segmentation models~\cite{entityseg,knet,maskformer,mask2former} reformulate image segmentation as binary mask extraction and mask classification problems. They typically update a series of learnable queries to represent all things and stuff in the input image. Then, the updated queries are utilized to conduct convolution on a feature map produced by the backbone and pixel-decoder to get binary masks for each object. At the same time, a classification head with fixed FC layers is added after each updated query to predict a class label from a predefined category set.

We pick Mask2Former~\cite{mask2former} as the base model. To make it compatible with the open-vocabulary setting, we remove its classification layer and project each initial query to a query embedding to match the text embeddings extracted by the CLIP text encoder. Thus, after normalization, we could get the logits for each category by calculating the cosine similarity. Since the values of cosine similarity are small, it is crucial to make the distribution sharper when utilizing the softmax function during training. Hence, we add a temperature parameter $\tau$ as 0.01 to amplify the logits.

We train our vanilla model on COCO~\cite{coco} using panoptic annotations. Following unified segmentation methods~\cite{mask2former,maskformer,knet}, we apply bipartite matching to assign one-on-one targets for each predicted query embedding, binary mask, and IOU score. We apply cross-entropy loss on the softmax normalized cosine similarity matrix to train the mask-text alignment. For the binary masks, we apply dice loss~\cite{vnet} and binary cross-entropy loss. More details refer to ~\cite{mask2former}.

\subsection{Leveraging  CLIP Visual Features}
\label{sec:CLIP embeddings}

Instead of training the query embeddings with large amount of data like \cite{openseg,Detic}, 
we investigate introducing the pretrained CLIP visual embeddings for better object recognition. Similarly, some works~\cite{simplebaselinezsg,zegformer} pass each masked proposal into the CLIP image encoder to extract the visual embedding. However, this strategy has the following drawbacks: first, it is extremely inefficient, especially when the object number is big; second, the masked region lacks context information, which is harmful for recognition.   

Conducting mask-pooling on the CLIP features seems a straightforward solution. However, CLIP image encoder uses an attention-pooling layer to reduce the spatial dimension and makes image-text alignment simultaneously. We use a Spatial Adapter to maintain its resolution. Concretely, we re-parameterize the linear transform layer in attention-pooling as $1\times1$ convolution to project the feature map into language space.

Getting the CLIP visual features, on the one hand, we make information exchange with the segmentation model via using the CLIP features to enhance the query embeddings through cross attention. On the other hand, we adopt Mask Pooling which utilizes the binary masks to pool them into CLIP embeddings. These embeddings contain the generalizable representation for each proposal.

\subsection{Embedding Modulation} 
\label{sec:ecm}
Both the query embeddings and the CLIP embeddings could be utilized for recognition. We analyze that, as the query embeddings are trained, they have advantages in predicting in-domain categories, whereas the CLIP embeddings have priorities for unfamiliar novel categories. Therefore, we develop Embedding Modulation that takes advantage of those two embeddings and enables adequate embedding enhancement and information exchange for them, thus advancing the recognition ability and making OPSNet omnipotent for generalized domains and data-efficient for training. The Embedding Modulation contains two steps.

\vspace{-4mm}
\paragraph{Embedding Fusion.}
We first use the CLIP text encoder to extract text embeddings for the $\text N$ category names of the training data, and for the $\text M$ names of the predicting concept set. Then, we calculate a cosine similarity matrix $\mathbf{H}^{\text{M} \times \text{N}}$ between the two embeddings. Afterward, we calculate a domain similarity coefficient $\text s$ for the target concept set as $\text{s} = \frac{1}{\text{M}} \sum_{i}  \emph{max}_{j}(\mathbf{H}_{i,j})$, which means that for each category in the predicting set, we find its nearest neighbor in the training set by calculating the cosine similarity, and then they are averaged to calculate the domain similarity.

With this domain similarity, we fuse the query embeddings $\mathbf{E}_{q}$ and the CLIP embeddings $\mathbf{E}_{c}$ to get the modulated embeddings 
$\mathbf{E}_{m} = \mathbf{E}_{q} + \alpha\cdot(1-\text{s})\cdot\mathbf{E}_{c}$. 
The principle is, the fusion ratio between the two embeddings is controlled by the domain similarity $\text s$, as well as a $\alpha$ which is 10 as default.

\vspace{-3mm}
\paragraph{Logits Debiasing.}
With the modulated embeddings, we get the category logits by computing the cosine similarity between the  modulated embeddings and the text embeddings of category names. We denote the logits of the \textit{i}-th category as $\ve{z}_i$.
Inspired by \cite{logitadjustment}, which uses frequency statistics to adjust the logits for long-tail recognition, in this work, we use the concept similarity  to debias the logits, thus balancing seen and unseen categories as $\hat{\ve{z}}_i = \ve{z}_i ~/~ ( \emph{max}_{j}( \mathbf{H}_{i,j})_i)^\beta $, where $\beta$ is a coefficient controls the adjustment intensity.
The equation means that, for the \textit{i}-th category, we find the most similar category in the training set and use this class similarity to adjust the logits. In this way, the bias towards seen categories could be alleviated smoothly. The default value of $\beta$ is 0.5.

\subsection{Additional Improvements}
\label{sec:additional}
The framework above is already able to make open-vocabulary predictions. In this section, we propose two additional improvements to push the boundary of OPSNet.      

\vspace{-5mm}
\paragraph{Mask Filtering.}
Leveraging the CLIP embeddings for modulation is crucial for improving the generalization ability, but it also raises a problem: the query-based segmentation methods~\cite{maskformer,mask2former} rely on the classification predictions to filter invalid proposals to get the panoptic results. Concretely, they add an additional background class and assign all unmatched proposals as background in Hungarian matching. Thus, they could filter invalid proposals during inference without NMS. Without this filtering process, there would be multiple duplicate or low-quality masks. 

However, the CLIP embeddings are not trained with this intention. Thus, we should either adapt the CLIP embeddings for background filtering or seek other solutions. To address the issue, we design Mask Filtering to filter invalid proposals according to the estimated mask quality. We add an IoU head with one linear layer to the segmentation model after the updated queries. It learns to regress the mask IoU between each predicted binary mask and the corresponding ground truth. For unmatched or duplicated proposals, it learns to regress to zero. We use an $L_2$-loss to train the IoU head and utilize the predicted IoU scores to rank and filter segmentation masks during testing. As the IoU is not relevant to the category label, it could naturally be generalized to unseen classes. This modification enables our model the ability to detect and segment more novel objects, which serves as the essential step towards open vocabulary.

\vspace{-4mm}
\paragraph{Decoupled Supervision.} 
Common segmentation datasets like \cite{coco,ade20k,cityscape,mapillaryvistas} contain less than 200 classes, 
but image classification datasets cover far more categories. Therefore, it is natural to explore the potential of classification datasets. Some previous works~\cite{Detic,openseg} attempt to use of image-level supervision. However, the strategy of Detic~\cite{Detic} is not extendable for multi-label supervision;  OpenSeg~\cite{openseg} designs a contrastive loss requiring a very large batch size and memory, which is hard to follow. Besides, they only supervise the classification but give no constraints for segmentation.   
In this situation, we develop Decoupled Supervision, a superior paradigm that utilizes image-level labels to improve the generalization ability, and digging supervisions from the predictions themselves for training masks. We denote this advanced version as OPSNet$^+$.

For a classification dataset with C categories, we extract the text embeddings $\mathbf{T}^{\text{C} \times \text{D}}$ with $\text{D}$ dimensions.
For a specific image with $\text c$ annotated object labels, assuming that OPSNet gives K predicted binary masks $\mathbf{M}^{\text{K} \times \text{H} \times \text{W}}$ with spatial dimension $\text{H}\times \text{W}$, modulated embeddings $\mathbf{E}_m^{\text{K} \times \text{D}}$, and IoU scores $\mathbf{U}^{\text{K} \times 1}$.     
We first remove the invalid predictions if their IoU scores are lower than a threshold, resulting in $\text J$ valid predictions. At the same time, we pick the embeddings $\mathbf{E}_m^{\text{J} \times \text{D}}$ for each valid prediction. 
We compute the cosine similarity of this selected  embeddings and the text embeddings $\mathbf{T}^{\text{C} \times \text{D}}$  and obtain a similarity matrix $\mathbf{S}^{\text{J}\times \text{C}}$.

We normalize each row~(the first dimension) of  $\mathbf{S}^{\text{J}\times\text{C}}$ using a softmax function $\delta$. Afterwards, we select the max value along the first dimension of $\delta(\mathbf{S}^{\text{J}\times \text{C}} )$, and select the columns~(the second dimension) for the  c annotated categories. We note this column selection operations as $\mathbbm{1}_{j\in{\mathbbm{R}^c}}$. The matching loss could be formulated as:

\begin{equation}
    \mathcal{L}_{match} = 1 - \frac{1}{ \text{c} } \sum\limits_{j=1}^\text{c} \emph{max}_i  (\delta( \mathbf{S}_{i,j} )) \mathbbm{1}_{j\in{\mathbbm{R}^c}} 
    \label{eqn:matching loss}
\end{equation}

This loss encourages the model to predict at least one matched embeddings for each image-level label. The model will not be penalized if it exists multiple masks for one category, or if there exists missing GT labels.

Although without mask annotations, the layout of panoptic masks could be regarded as supervision. As we expect the predicted masks to fill the full image, and do not overlap with each other, the summation of all predicted masks could be formulated as a constraint.
Concretely, we normalize all the K predicted masks using the Sigmoid function $\sigma$ and add all K masks to one channel. We encourage each pixel of the mask to get close to one, and propose a sum loss as: 
\begin{equation}
    \mathcal{L}_{sum} = || 1 - \sum_{k=1}^{\text{K}} (\sigma(\mathbf{M}_{k,i,j}))||_2
    \label{eqn:sum loss}
\end{equation}

When introducing ImageNet for training,  we  add $\mathcal{L}_{match}$ and $\mathcal{L}_{sum}$ with weights of $1.0$ and $0.4$.
\section{Experiments} \label{sec:exp}
\paragraph{Implementation details.} We adopt Mask2Former~\cite{mask2former} as our segmentation model, 
and choose the ResNet-50~\cite{resnet} version CLIP~\cite{clip} for visual-language alignment, where the image and text are encoded as 1024-dimension feature vectors. Compared with Mask2Former, the additional computation burden of CLIP is acceptable as we choose the smallest version of CLIP and do not compute the gradient.  When using the Swin-L backbone for Mask2Former, with an input size of 640, the FLOPs and Params of Mask2Former and OPSNet are 403G/485G and 215M/242M. As we pass CLIP only once, our FLOPs is significantly smaller than \cite{zegformer,simplebaselinezsg}, which feed each proposal into CLIP.

\vspace{-2mm}
\paragraph{Training configurations.} In the basic setting, we train on the COCO~\cite{coco} panoptic segmentation training set. The hyper-parameters follow Mask2Former. The training procedure lasts 50 epochs with AdamW~\cite{adamw} optimizer. The initial learning rate (LR) is 0.0001, and it is decayed with the ratio of 0.1 at the 0.9 and 0.95 fractions of the total steps.

For the advanced version with extra image-level labels, we mix the classification data with COCO panoptic segmentation data. The re-annotated ImageNet~\cite{imagenetpp} is utilized where correct multi-label annotations are included. We use the validation split for simplicity, which covers 1K categories and contains 50 images for each category. When calculating the losses, the category names from COCO and ImageNet are treated separately.
We finetune OPSNet for 80K iterations ($\sim$ 5 epochs). The initial LR is 0.0001 and multiplied by 0.1 at the 50K iteration.

\vspace{-2mm}
\paragraph{Evaluation and metrics.} We evaluate OPSNet for both open-vocabulary and closed-world settings.
We evaluate the open-vocabulary ability by conducting cross-dataset validation for panoptic segmentation on ADE20K~\cite{ade20k}, and Cityscapes~\cite{cityscape}. To evaluate the closed-world ability, we also compare OPSNet with SOTAs on COCO panoptic segmentation. 
We report the overall PQ~(Panoptic Quality), the PQ for things and stuff, the SQ~(Segmentation Quality), and the RQ~(Recognition Quality).
Then, we report the mIoU~(mean Intersection over Union) for semantic segmentation on ADE20K~\cite{ade20k} and Pascal Context~\cite{pascalcontext} to compare with previous works. Afterward, we use the large concept set of ImageNet-21K~\cite{imagenet} and give qualitative results for open-vocabulary prediction and hierarchical prediction.

\subsection{Roadmap to Open-vocabulary Segmentation}
\begin{table*}[t]
\begin{center}
\scalebox{0.8}{
\begin{tabular}{ll|ccccc|ccccc}
\toprule[1pt]
\multicolumn{2}{l|}{} & \multicolumn{5}{c|}{COCO}
& \multicolumn{5}{c}{ADE20K} \\
 & Method & PQ & PQ$^{th}$ & PQ$^{st}$ & SQ & RQ & PQ & PQ$^{th}$ & PQ$^{st}$ & SQ & RQ \\
\hline
1 & Mask2Former~\cite{mask2former} & 51.9 & 57.7 & 43.0 & 83.1  & 61.6 & - & - & - & - & - \\

2 & CAG-Seg + CLIP Embeds  & 12.5 & 17.7  & 4.6 & 68.1 & 15.3 & 4.9 & 5.2 & 4.2 & 45.5 & 6.2  \\
3 & CAG-Seg + CLIP Embeds + Mask Filter & 22.7 & 26.9 & 16.3  & 82.1 & 26.7 & 10.7 & 9.5 & 13.3 & 66.6 & 13.1  \\
4 & CAG-Seg + Query Embeds & 51.5 & 57.3 & 42.8 & 83.2 & 61.1 & 13.6 & 11.3  & 18.0 & 29.8 & 16.8 \\
5 & CAG-Seg + Query Embeds + Mask Filter & 51.9 & 57.4 & 43.4 & 83.3 &  61.5 & 14.5 & 12.4  & 19.3 & 37.7 & 17.6  \\

6 & CAG-Seg + Query Embeds$^{\dagger}$ + Mask Filter & 52.4 & 58.0 & 44.0  & 83.5 & 62.1 & 14.6 & 13.2  & 17.6 & 33.8 & 17.1 \\

\rowcolor{gray!20} 
7 & \textbf{OPSNet} (CAG-Seg + Modulated Embeds + Mask Filter )  & \bf 52.4 & \bf 58.0 &  \bf44.0  &  \bf83.5 &  \bf62.1 &  \bf17.7 & \bf 15.6  &  \bf21.9 &  \bf54.9 & \bf21.6 \\

\bottomrule
\end{tabular}
}
\end{center}

\vspace{-2mm}
\caption{ Ablation study for the roadmap towards open-world panoptic segmentation. All experiments use ResNet-50 backbone, and are trained on COCO for 50 epochs. `CAG-Seg' denotes the class-agnostic segmentation model. `Query Embeds$^{\dagger}$' means adopting the cross attention layer to gather information from the CLIP features.   }
\label{tab:crossade}
\vspace{-2mm}
\end{table*}

\begin{table*}[t]
\begin{center}
\scalebox{0.75}{
\begin{tabular}{ll|ccccc|ccccc|ccccc}
\toprule[1pt]
\multicolumn{2}{l|}{} & \multicolumn{5}{c|}{COCO}  & \multicolumn{5}{c|}{ADE20K}
& \multicolumn{5}{c}{CityScapes} \\
Method & Backbone  & PQ & PQ$^{th}$ & PQ$^{st}$ & SQ & RQ  & PQ & PQ$^{th}$ & PQ$^{st}$ & SQ & RQ & PQ & PQ$^{th}$ & PQ$^{st}$ & SQ & RQ \\
\hline
MaskCLIP-Base~\cite{MaskCLIP} & ResNet-50  &  - & - &  - & - & -  & 9.6 & 8.9 & 10.9 & 62.5 &  12.6 & - & - & - & - & -\\
MaskCLIP-RCNN~\cite{MaskCLIP} & ResNet-50  &  - & - &  - & - & -  & 12.9 & 11.2 & 16.1 & 64.0 &  16.8 & - & - & - & - & -\\
MaskCLIP-Full~\cite{MaskCLIP} & ResNet-50  &  30.9 & 34.8 &  25.2 & - & -  & 15.1 & 13.5 & 18.3 & 70.5 &  19.2 & - & - & - & - & -\\
\hline
OPSNet & ResNet-50  &  52.4 & 58.0 &  44.0 & 83.5 & 62.1  & 17.7 & 15.6 & 21.9 & 54.9 &  21.6 & 37.8 & 35.5 & 39.5	& 64.2  &  45.8  \\

OPSNet & ResNet-101  & 53.9 & 59.6 & 45.3 & 83.6 & 63.7 & 18.2 & 16.0 &22.6 & 52.1  & 22.0 & 40.2 & 37.0 & 42.5  &	64.3 & 48.5\\

OPSNet & Swin-S &  54.8 & 60.5 & 46.2 & 83.7 & 64.8 & 	18.3 &  16.8 &  21.3 & 59.4 & 22.3 & 41.1 & 36.0  &44.8 & 66.9 &49.6 \\
\rowcolor{gray!20} 
OPSNet & Swin-L$^\dagger$   & 57.9 & 64.1 & 48.5 & 84.1 & 68.2 &19.0 & 16.6 & 23.8 & 52.4 & 23.0 & 41.5 & 36.9 & 44.8 & 67.5 & 50.0	\\
\bottomrule
\end{tabular}
}
\end{center}
\vspace{-2mm}
\caption{Open-vocabulary panoptic segmentation on different datasets with different backbones. All models are trained on COCO. `Swin-L$^\dagger$' denotes pre-trained on ImageNet-21K. Following \cite{mask2former}, we train the Swin-L$^\dagger$ version 100 epochs, and 50 epochs for other versions.}
\label{tab:panoptic}
\vspace{-2mm}
\end{table*}
\begin{table}[t]
\begin{center}
\scalebox{0.63}{
\begin{tabular}{lc|ccc|ccc}
\toprule[1pt]
\multicolumn{2}{l|}{} & \multicolumn{3}{c|}{ADE20K} & \multicolumn{3}{c}{COCO} \\
Method &  Backbone  & PQ & PQ$^{th}$ & PQ$^{st}$ & PQ & PQ$^{th}$ & PQ$^{st}$ \\
\hline
OPSNet & ResNet-50 & 17.7 & 15.6 & 21.9 & 52.4 & 58.0 & 44.0 \\
+ Cls Sup &  ResNet-50   & 18.2 & 15.0  & 24.4 & 51.3 & 56.9  & 42.9 \\
\rowcolor{gray!20} 
+ Cls Sup + Mask Sup  &  ResNet-50   & 19.0 & 16.6  & 23.9 & 51.7 & 57.2  & 43.4 \\

\hline
\rowcolor{gray!20} 
+ Cls Sup + Mask Sup  &  Swin-L   & 20.5 & 18.5  & 24.5 & 56.2 & 61.7  & 47.7 \\
\bottomrule
\end{tabular}
}
\end{center}
\vspace{-2mm}
\caption{Ablations for Decoupled Supervision. We use ImageNet-Val for additional data to expand the training concepts.} 
\label{tab:imagesup}
\vspace{-2mm}
\end{table}

\begin{table}[t]
\begin{center}
\scalebox{0.63}{
\begin{tabular}{ll|ccc|ccc|c}
\toprule[1pt]
\multicolumn{2}{l|}{} & \multicolumn{3}{c|}{COCO} & \multicolumn{3}{c|}{ADE20K} & PC \\
Embedding &  Setting  & PQ & PQ$^{th}$ & PQ$^{st}$ & PQ & PQ$^{th}$ & PQ$^{st}$ & mIOU \\
\hline
\multirow{2}*{Single}   & CLIP   & 22.7 & 26.9 & 16.3  & 10.7 & 9.5 & 13.3  & 27.3\\
~   & Query   & 51.9 & 57.4 & 43.4 & 14.5 & 12.4 & 19.3 & 45.3\\
\hline
\multirow{4}*{Ensemble} & Query + 1$\times$CLIP  & 51.4 & 56.9 & 43.1 & 16.4 & 14.2 & 20.7 &  48.3\\
~ & Query + 2$\times$CLIP   & 50.1 & 55.3 & 42.2 & 17.7 & 15.7 &21.6 & 47.4\\
~ & Query + 3$\times$CLIP    & 47.9 & 52.8 & 40.5 & 18.1 & 16.1 & 21.9 & 46.0\\
~ & Query + 4$\times$CLIP    & 43.8 & 48.1 & 37.3 & 17.4 & 15.3 & 21.6 & 41.9\\
\hline
Modulation & EF & 52.4 & 58.0 & 44.0 & 16.9 & 15.8  & 19.0 & 49.7\\
\rowcolor{gray!20} 
Modulation & EF + LD & 52.4 & 58.0 & 44.0  & 17.7 & 15.6 & 21.9 & 50.2\\
\bottomrule
\end{tabular}
}
\end{center}
\vspace{-2mm}
\caption{Ablation study for Embedding Modulation. `EF' denotes embedding fusion, `LD' means logits debiasing, `PC' stands for Pascal Context dataset.}
\label{tab:modulation}
\end{table}
\begin{table}[t]
\begin{center}
\scalebox{0.7}{
\begin{tabular}{lc|ccc|ccc}
\toprule[1pt]
\multicolumn{2}{l|}{} & \multicolumn{3}{c|}{ADE20K} & \multicolumn{3}{c}{COCO} \\
Method &  Pass Times  & PQ & PQ$^{th}$ & PQ$^{st}$ & PQ & PQ$^{th}$ & PQ$^{st}$ \\
\hline
Masking  & $\times$ N   & 6.7 & 5.9 & 8.5 & 16.3 & 21.5 & 8.4 \\
Cropping &  $\times$ N  & 9.4 & 8.9 & 12.0 & 19.6 & 24.5 & 14.1 \\
\rowcolor{gray!20} 
Mask-Pooling & $\times$ 1  & 10.7 & 9.5 & 13.3 & 22.7 & 26.9 & 16.3 \\
\bottomrule
\end{tabular}
}
\end{center}
\vspace{-2mm}
\caption{ Different methods for extracting CLIP embeddings. N means the number of objects in the image.} 
\label{tab:clipembed}
\vspace{-1mm}
\end{table}

\definecolor{aliceblue}{rgb}{0.94, 0.97, 1.0}
\definecolor{babyblueeyes}{rgb}{0.63, 0.79, 0.95}
\definecolor{celadon}{rgb}{0.67, 0.88, 0.69}
\definecolor{lightblue}{rgb}{0.68, 0.85, 0.9}

\begin{table}[t]
\begin{center}
\scalebox{0.7}{
\begin{tabular}{l|c|c|c|c|c }
\toprule[1pt]
\diagbox{$\alpha$}{$\beta$}  & $w/o$ LD & 0.25  & 0.5  & 0.75 & 1.0  \\
\hline
$w/o$ EF  &  \cellcolor{celadon} 14.7 & 16.1 & 16.3 & 15.1 & 14.2 \\
5 & 16.3 &  \cellcolor{aliceblue} 16.8  &  \cellcolor{aliceblue} 16.6 & \cellcolor{aliceblue} 16.2 & \cellcolor{aliceblue} 15.3  \\
10  & 16.9 & \cellcolor{aliceblue} 17.5 &   \cellcolor{lightblue} 17.7  &  \cellcolor{aliceblue}  16.7 & \cellcolor{aliceblue}  16.1 \\
15  & 17.1  & \cellcolor{aliceblue} 16.8  &  \cellcolor{aliceblue} 17.2 & \cellcolor{lightblue} 17.9 &  \cellcolor{aliceblue} 17.3  \\
\bottomrule[1pt]
\end{tabular}
}
\end{center}
\vspace{-2mm}
\caption{Grid search for the $\alpha,\beta$ of Emedding Modulation. Results on ADE20K panoptic dataset are reported.}
\label{tab:grid}
\vspace{-2mm}
\end{table}

We introduce our roadmap for building an open-vocabulary segmentation model. We first describe the overall procedure for how to equip our vanilla solution to OPSNet step by step. Then, we dive into the details to analyze each of our exquisite components. Following CLIP and OpenSeg~\cite{openseg}, we report the cross-dataset results for evaluating the generalization ability of our model. 

Besides, we claim that keeping the performance on the training domain is also important. Therefore, we report the performance of both  ADE20K and COCO~(training domain) for the ablation studies.

\vspace{-2mm}
\paragraph{From vanilla solutions to OPSNet.}
In Table~\ref{tab:crossade}, we conduct experiments on COCO and ADE20K panoptic data step by step from vanilla solutions to OPSNet.

The closed-vocabulary method Mask2Former cannot directly evaluate other datasets due to the category conflicts. 
In row 2, we remove its classification head to make it predict  class-agnostic masks. Then, as introduced in Sec.~\ref{sec:CLIP embeddings}, we use these masks to pool the CLIP features to get CLIP embeddings, and use them for recognition. However, as explained in Sec.~\ref{sec:additional}, this modification would not be suitable if we still adopt the classification results to filter the proposals. Therefore, in row 3, we add Mask Filtering and observe significant performance improvements.
In rows 4 and 5, we show the performance of only using the query embeddings for recognition. Then, in row 6, we demonstrate that adding a cross-attention layer to gather the CLIP features would be helpful for learning query embeddings. Finally, in row 7, we add the Embedding Modulation for the full-version OPSNet, which shows a great gain in generalization.

The experimental results show that with the adequate embedding enhancement and the
information exchange between CLIP and the segmentation model.  Even only trained on COCO, OPSNet archives great performance on both COCO and ADE20K datasets.

\paragraph{More data with Decoupled Supervision.} As introduced in Sec.~\ref{sec:additional}, we develop a superior training paradigm that utilizes image-level labels. In Table~\ref{tab:imagesup}, beside using COCO annotations, we further improve the generalization ability of OPSNet by introducing 50,000 images from the relabeled version of ImageNet-Val~\cite{imagenetpp}.
We first verify the effectiveness of each decoupled supervision. Then we report the performance of different backbones. When more training categories are introduced, the cross-dataset ability of OPSNet improves significantly, as indicated by the exceptional performance of OPSNet$^+$.

\begin{figure*}[t]
\newcommand{\image}{\includegraphics[width=0.97\linewidth]}
\centering 
\image{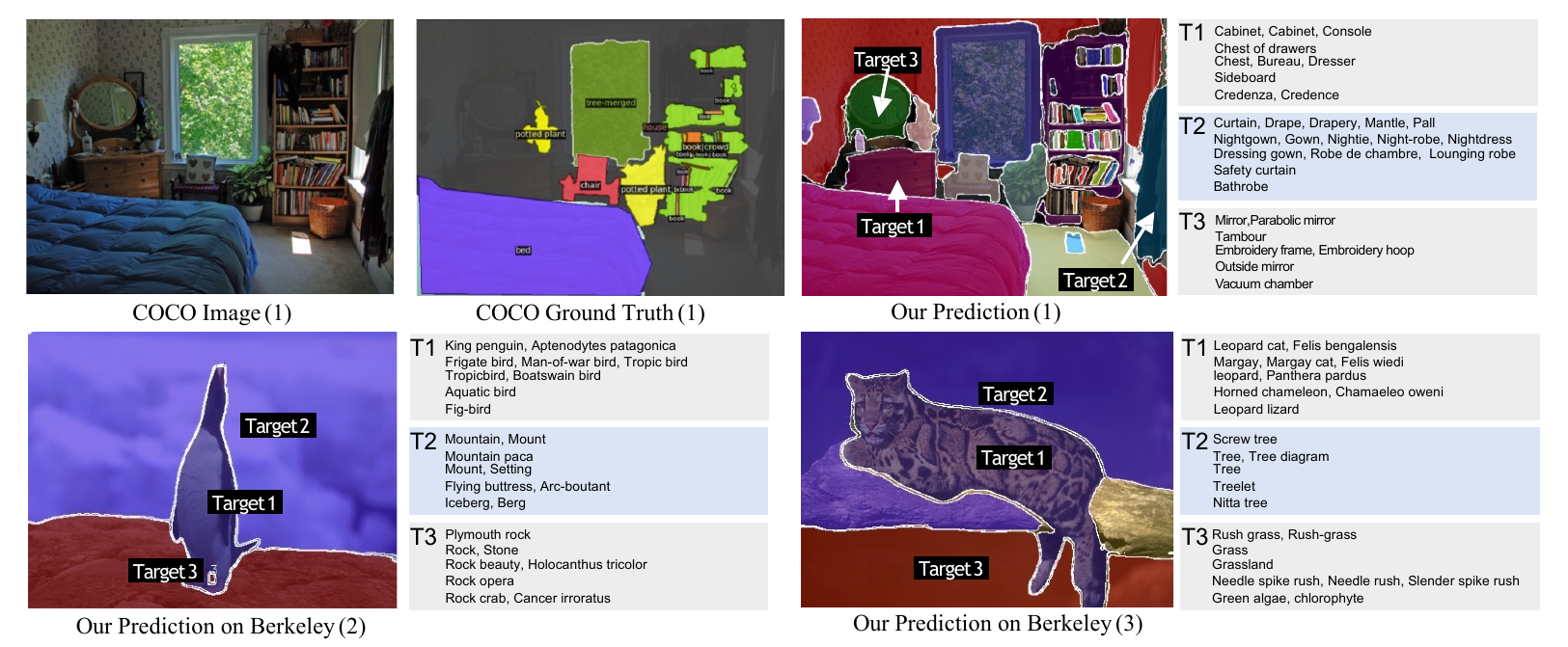}
\caption{Illustrations of open-vocabulary image segmentation. We choose the 21K categories of ImageNet as our prediction set. We display five proposals  with the highest confidence. OPSNet could make predictions for categories that are not included in COCO.}
\vspace{-3mm}
\label{fig:demo1}
\end{figure*}

\vspace{-2mm}
\paragraph{Analysis for CLIP embedding extraction.} In Table~\ref{tab:clipembed}, we verify the priority of our Spatial-Adapter and Mask Pooling using pure CLIP embeddings. This design shows better recognition ability and efficiency.

\vspace{-2mm}
\paragraph{Analysis for Embedding Modulation.} We give an in-depth analysis of the modulation mechanism. In Table~\ref{tab:modulation}, we report the results of the naive ensemble strategies between the query embeddings and the CLIP embeddings. We simply add these two embeddings with different ratios, and surprisingly find this straightforward method quite effective. However, we observe that the best ratio is different for each target dataset, a specific ratio would be beneficial for certain datasets but harmful for others. Our modulation strategy controls this ratio according to the domain similarity between the training and target sets and debias the final logits using the categorical similarity, which shows a strong balance across different domains.

In Table~\ref{tab:grid}, we carry out grid search for the coefficient $\alpha$ and $\beta$ which control the modulation intensity. The results show the robustness of proposed method.  

\begin{table}[t]
\begin{center}
\scalebox{0.72}{
\begin{tabular}{ll|ccc|ccc}
\toprule[1pt]
\multicolumn{2}{l|}{} & \multicolumn{3}{c|}{COCO} & \multicolumn{3}{c}{ADE20K}  \\
Embedding & Filter  & PQ & PQ$^{th}$ & PQ$^{st}$ & PQ & PQ$^{th}$ & PQ$^{st}$ \\
\hline
\multirow{3}*{Query} & Cls & 51.5 & 57.3 & 42.8 & 13.6 & 11.3 & 18.0 \\
~ & IoU   & 50.2 & 56.0  & 41.7 & 14.3 & 12.3  & 18.4 \\
\rowcolor{gray!20} 
~ & Cls$\cdot$IoU & 51.9 & 57.4 &  43.4  & 14.5 & 12.4 & 19.3  \\
\hline
{CLIP} & Cls & 12.5 & 17.7 & 4.6 & 4.9 & 5.2 & 4.2 \\

\rowcolor{gray!20} 
~ & Cls$\cdot$IoU & 22.7 & 26.9 & 16.3  & 10.7 & 9.5 & 13.3   \\
\hline
\multirow{3}*{Modulated} & Cls   & 50.0 & 55.4  & 41.9 & 14.9 & 40.6 & 18.1 \\
~ & IoU   & 50.3 & 56.1  & 41.6 & 16.0 & 14.3 &19.4 \\
\rowcolor{gray!20} 
~ & Cls$\cdot$ IoU  & 51.4 & 56.9 & 43.1 & 16.4 & 49.2 & 19.8 \\
\bottomrule
\end{tabular}
}
\end{center}
\vspace{-2mm}

\caption{Ablation study for Mask Filtering. `Cls$\cdot$ IoU' means the multiplication of the classification score and IoU score.  }
\label{tab:iou}
\vspace{-1mm}
\end{table}
\begin{table}[t]
\small
\vspace{-5pt}
\begin{center}
\scalebox{0.72}
{
\begin{tabular}{lccccc }
\toprule[1pt]
Method & Backbone & Training Data &  ADE  & PC  & COCO \\
\midrule
ALIGN ~\cite{ALIGN,openseg} & Efficient-B7 & Classification Data  & 9.7 & 18.5 & 15.6 \\

ALIGN$^+$~\cite{openseg} & Efficient-B7 & COCO  & 12.9 &  22.4 & 17.9 \\

LSeg$^+$~\cite{lseg,openseg}  & ResNet-101  & COCO & 18.0 & 46.5 & 55.1 \\
SimBase~\cite{simplebaselinezsg} &  ResNet-101  & COCO & 20.5  & 47.7  & - \\
\rowcolor{gray!20} 
OPSNet  & ResNet-101  & COCO & \bf 21.7 & \bf 52.2 & \bf 55.2 \\

\hline
OpenSeg~\cite{openseg}  & ResNet-101  & COCO + Caption~(600K) &  17.5 & 40.1 & -  \\

OpenSeg~\cite{openseg}  & Efficient-B7  & COCO + Caption~(600K) &  24.8 & 45.9 & 38.1 \\
\rowcolor{gray!20} 
OPSNet$^+$  & ResNet-101  & COCO + ImageNet~(50K) & 24.5 & 54.3 & 61.4 \\
\rowcolor{gray!20} 
OPSNet$^+$  & Swin-L$^\dagger$  & COCO + ImageNet~(50K) & \bf 25.4 & \bf 57.5 & \bf 64.8 \\

\bottomrule[1pt]
\end{tabular}
}
\end{center}
\vspace{-2mm}

\caption{Open-vocabulary semantic segmentation. 
The results for `ALIGN', `ALIGN$^+$', `LSeg$^+$' are all the modified versions introduced in OpenSeg.}
\vspace{-5mm}
\label{tab:openseg}
\end{table}

\begin{figure}[t]
\newcommand{\image}{\includegraphics[width=0.28\columnwidth]}
\centering 
\tabcolsep=0.05cm
\renewcommand{\arraystretch}{0.06}
\hspace{-3mm}
\begin{tabular}{ccc}
\vspace{1mm}
\image{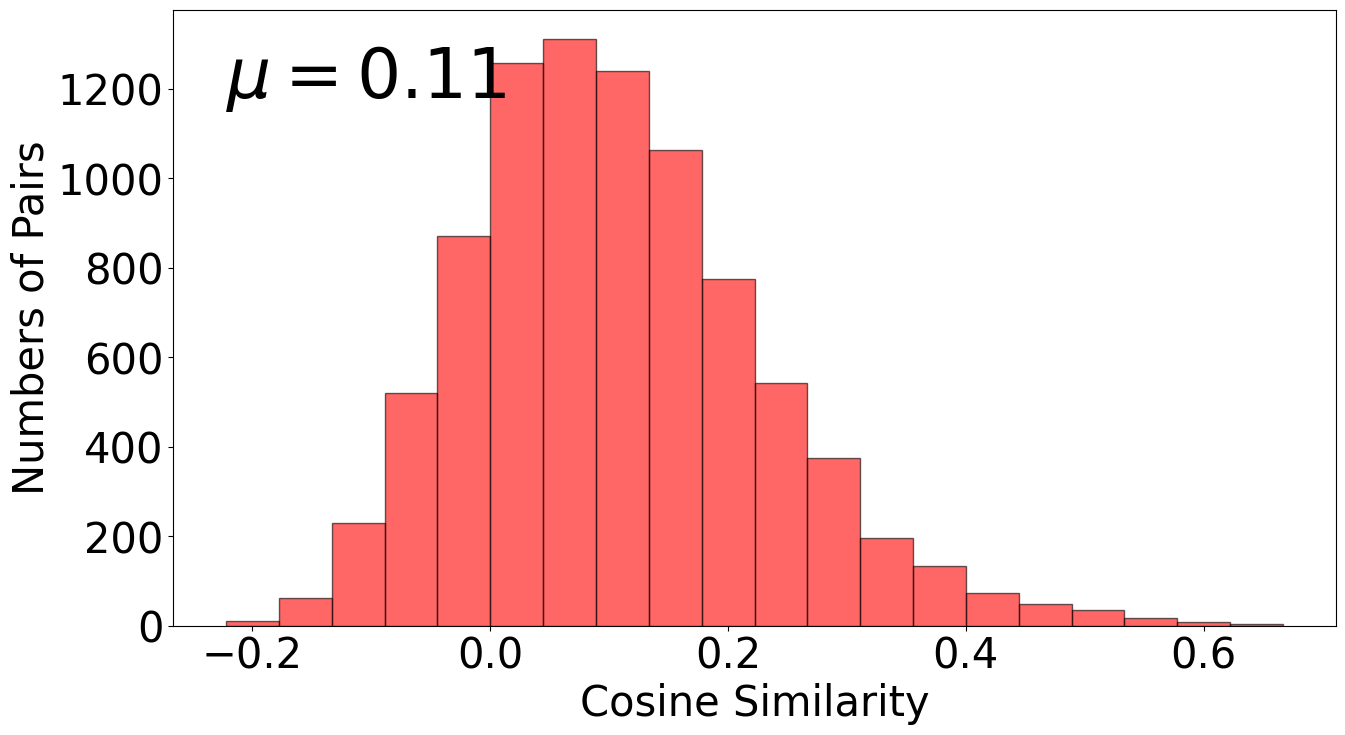} &
\image{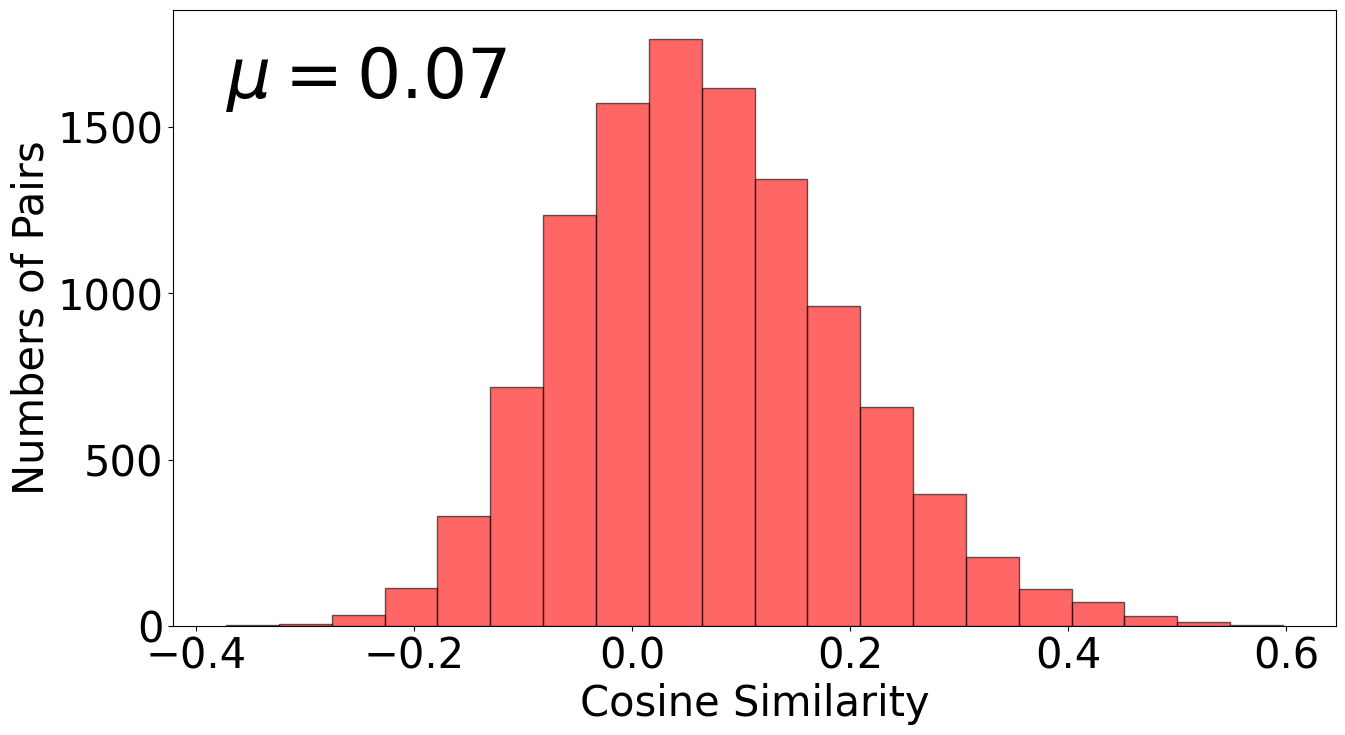} &
\image{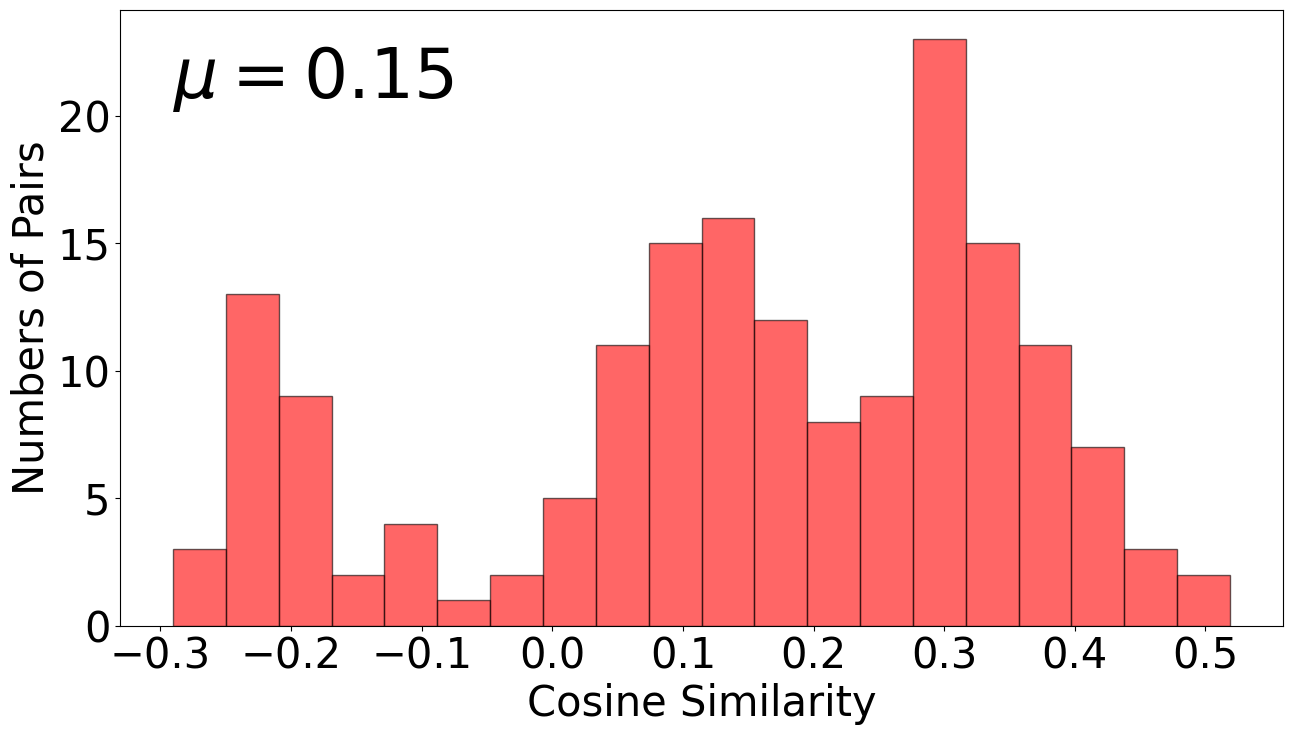} \\
\vspace{1mm}
{ \tiny (a)~COCO Protos } & {\tiny (b)~ADE20K Protos} & {\tiny (c)~CityScapes Protos} \\
\vspace{1mm}
\image{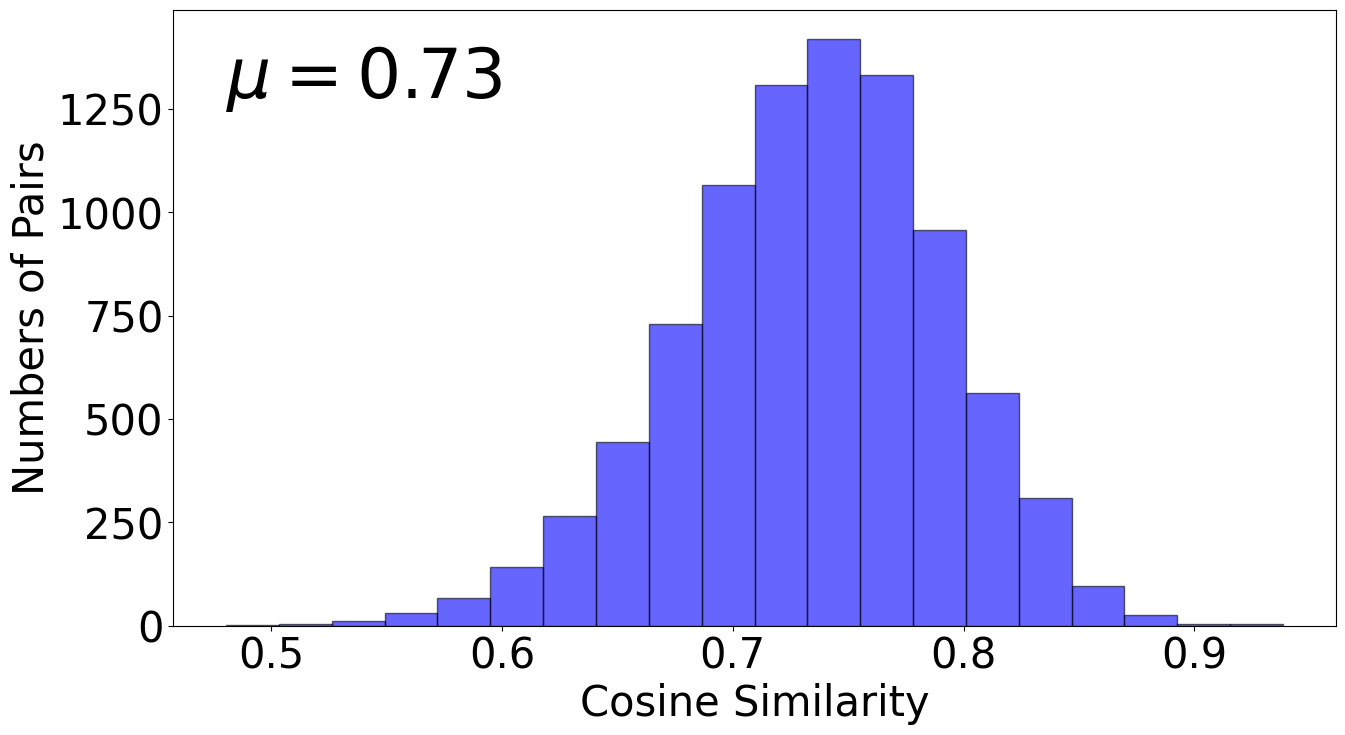} &
\image{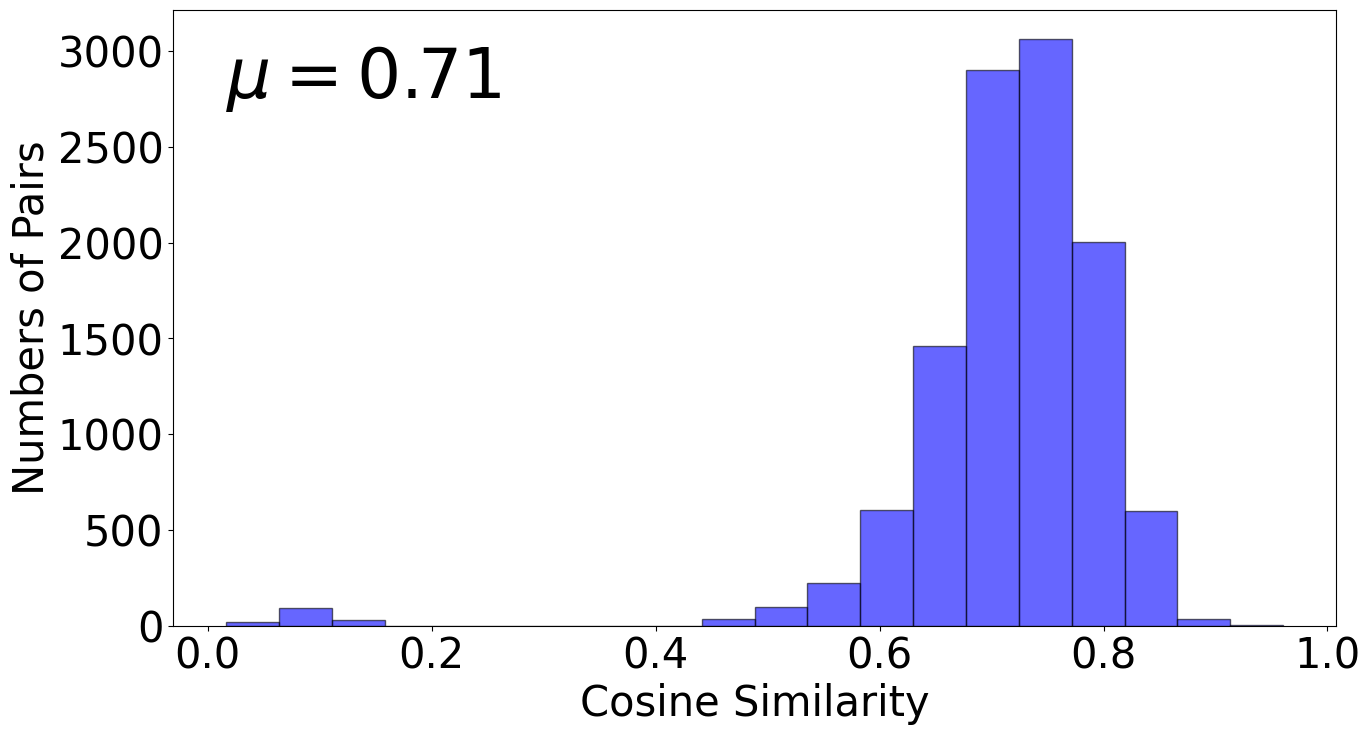} &
\image{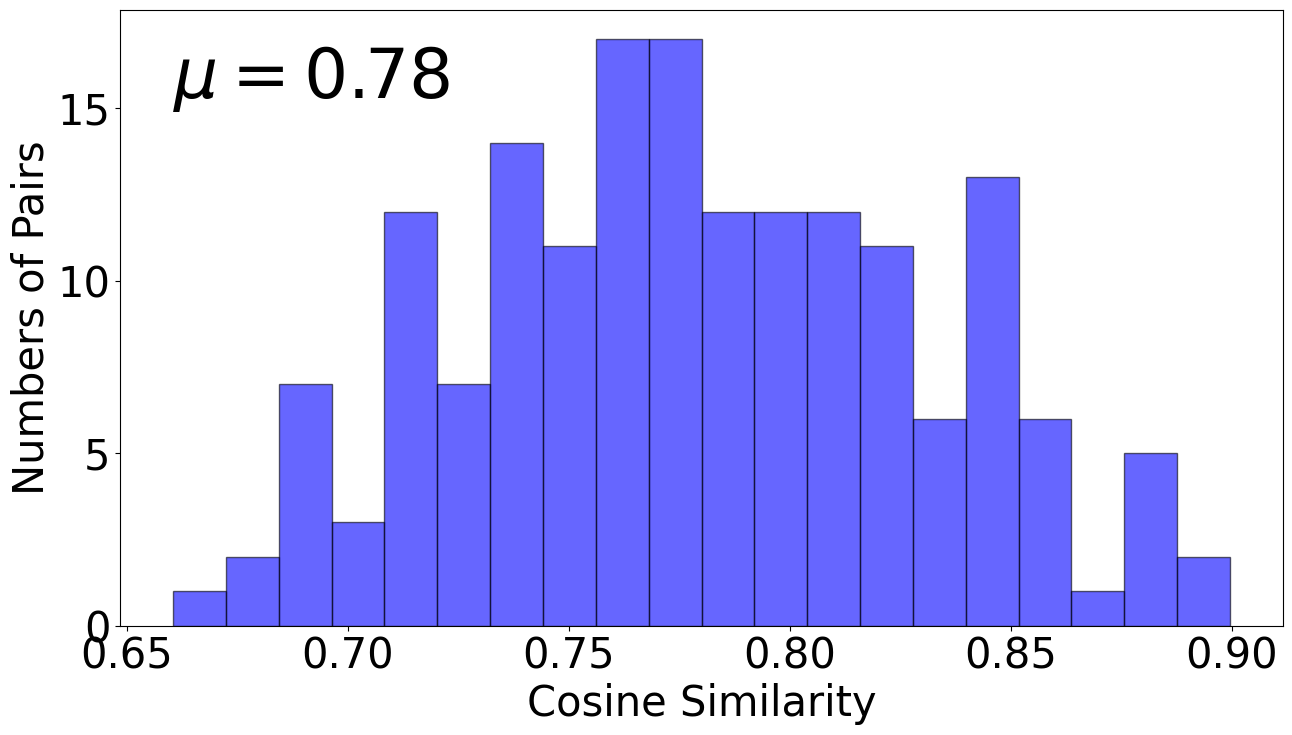} \\
{ \tiny (d)~COCO CLIP embeds } & {\tiny (e)~ADE20K CLIP embeds} & {\tiny (f)~CityScapes CLIP embeds} \\
\end{tabular}
\caption{The distributions of the pairwise cosine similarities among categories for the trained prototypes and  the CLIP text embeddings on different datasets. The mean value is noted as $\mu$. }
\vspace{-2mm}
\label{fig:sim}
\end{figure}

\vspace{-4mm}
\paragraph{Analysis for Mask Filtering.} First, to demonstrate the gap between closed-vocabulary  and open-vocabulary settings, in Fig.~\ref{fig:sim}, we compare the cosine similarity distribution between the trained class prototypes~(weights of the last FC layer) of Mask2Former and the CLIP text embeddings that used by OPSNet. We find the text embeddings are much less discriminative than the trained class prototypes, and the similarity distribution text embeddings vary for different datasets. Thus, the classification score of OPSNet would not be as indicative as the original Mask2Former to rank the predicted masks, which supports the claims in Sec.~\ref{sec:additional}.

In Table~\ref{tab:iou}, we conduct ablation studies with different visual embeddings. The three blocks correspond to the CLIP, query, and modulated embeddings respectively. The results show that an IoU score could notably improve performance especially when CLIP embeddings are introduced. 

\begin{figure*}[t]
\newcommand{\image}{\includegraphics[width=1.98\columnwidth]}
\centering 
\image{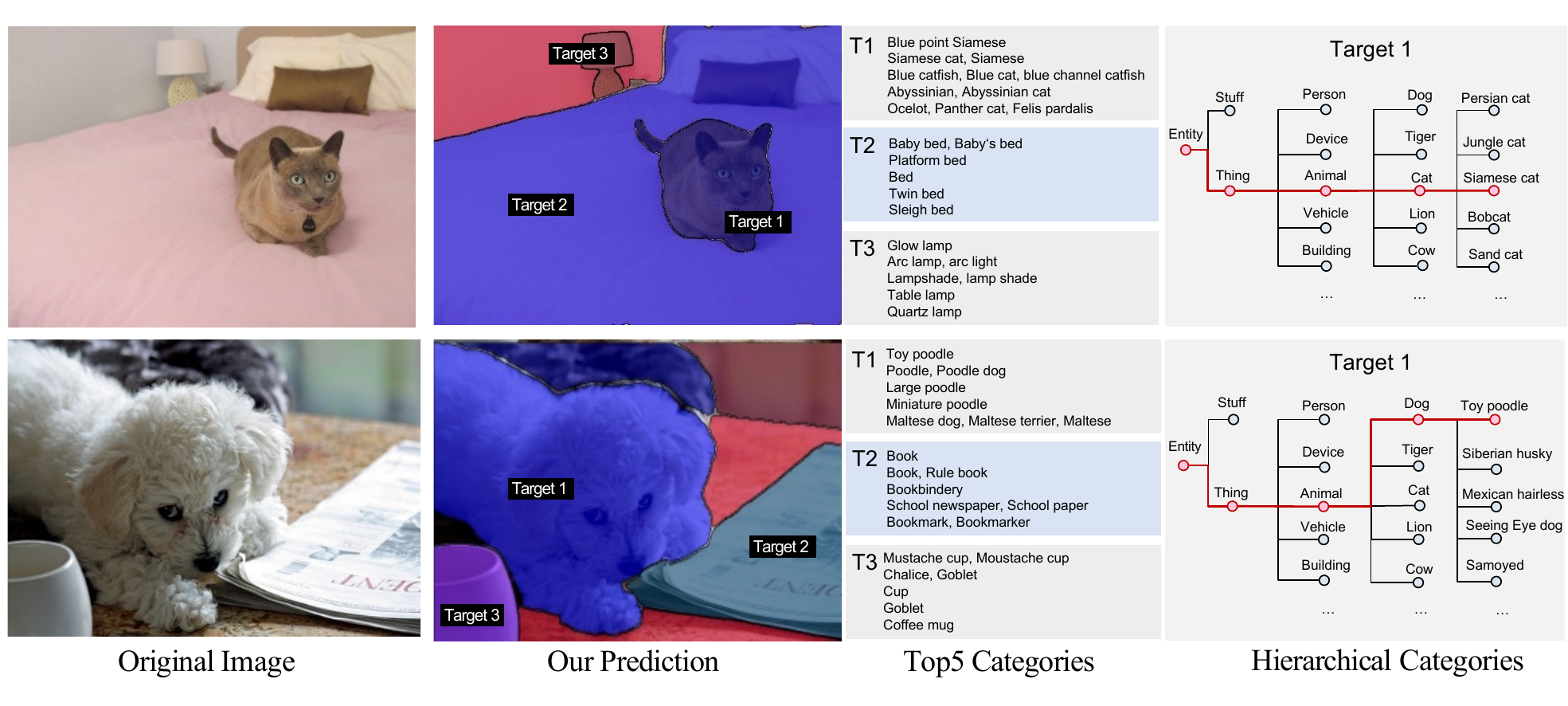} 
\vspace{-3mm}
\caption{Demonstrations for open-vocabulary image segmentation with hierarchical categories.}
\vspace{-4mm}
\label{fig:hie}
\end{figure*}

\vspace{-4mm}
\subsection{Cross-dataset Validation }
To evaluate the generalization ability of the proposed OPSNet, we conduct cross-dataset validation.

\vspace{-4mm}
\paragraph{Open-vocabulary panoptic segmentation.} In Table~\ref{tab:panoptic}, we report the results on three different panoptic segmentation datasets. 
OPSNet shows significant superiority over MaskCLIP~\cite{MaskCLIP} on both COCO and ADE20K, which verifies our omnipotent for general domains.

\vspace{-5mm}
\paragraph{Open-vocabulary semantic segmentation.} 
Some previous works~\cite{openseg,lseg,zegformer,simplebaselinezsg} explore open-vocabulary semantic segmentation. In Table~\ref{tab:openseg}, we make comparisons with them by merging our panoptic predictions into semantic results according to the predicted categories. 

Among previous methods, OpenSeg~\cite{openseg} is the most representative one. Here we emphasize our differences with OpenSeg: 1)~OpenSeg could only conduct semantic segmentation, as it does not deal with duplicated or overlapped masks. However, we develop Mask Filtering to remove the invalid predictions, thus maintaining the instance-level information. 2)~OpenSeg completely retrains the mask-text alignment, thus requiring a vast amount of training data. In contrast, Embedding Modulation efficiently utilizes features extracted by the CLIP image encoder, which makes our model data-efficient but effective.

OPSNet demonstrates superior results on all these datasets. Compared with OpenSeg, our model shows superiority using much fewer training samples. Besides, although OpenSeg reaches great cross-dataset ability, its performance on COCO is poor. In contrast, OPSNet keeps a strong performance in the training domain~(COCO), which is also important for a universal solution.

\begin{table}[t]
\small
\begin{center}
\scalebox{0.8}
{
\begin{threeparttable}
\begin{tabular}{l|ccccc }
\toprule[1pt]
Method & Backbone   & Epochs &  PQ  & PQ$^{th}$  &  PQ$^{st}$  \\
\hline
Max-DeepLab~\cite{maxdeeplab} & Max-L  & 216 & 51.1 &  57.0 & 42.2 \\
MaskFormer~\cite{maskformer} & Swin-L$^\dagger$ & 300 & 52.7 & 58.5 & 44.0 \\
Panoptic Segformer~\cite{panopticsegformer} & PVTv2-B5 & 50 & 54.1 & 60.4 & 44.6 \\
K-Net~\cite{knet} &  Swin-L$^\dagger$  & 36 & 54.6 & 60.2 & 46.0\\
\hline
\multirow{4}*{Mask2Former~\cite{mask2former}} & ResNet-50 & 50  & 51.9 & 57.7 & 43.0 \\
~ & ResNet-101 & 50  & 52.6 & 58.5 &  43.7 \\
~ & Swin-L$^\dagger$ & 100 & 57.8 & 64.2 & 48.1 \\
\hline
\multirow{4}*{OPSNet} & ResNet-50  & 50 & 52.4 & 58.0 &  44.0\\
~ & ResNet-101  & 50 & 53.9 & 59.6 & 45.3  \\
~ & Swin-L$^\dagger$ & 100 &  57.9 & 64.1 & 48.5 \\
\bottomrule[1pt]
\end{tabular}
\end{threeparttable}
}
\end{center}
\vspace{-2mm}
\caption{Closed-vocabulary panoptic segmentation on COCO validation set. Swin-L$^\dagger$ denotes pre-trained on ImageNet-21K.}
\label{tab:sotacocopanoptic}
\vspace{-5mm}
\end{table}

\vspace{-1mm}
\subsection{Closed-vocabulary Performance}
\vspace{-1mm}
We consider maintaining a competitive performance on the classical closed-world datasets is also important for a omnipotent solution. Therefore, in Table~\ref{tab:sotacocopanoptic}, we compare the proposed OPSNet with the current best methods for COCO panoptic segmentation. OPSNet gets better performance than our base model Mask2Former, and shows competitive results compared with SOTA methods.

\subsection{Generation to Broader Object Category}
\paragraph{Prediction with 21K concepts.} We use the categories of ImageNet-21K~\cite{imagenet} to describe the segmented targets. This large scope of words could roughly cover all common objects in everyday life. As illustrated in Fig.~\ref{fig:demo1}, we display the top-5 category predictions for several segmented masks.

The first row shows examples in COCO. The ground truth annotations ignore the objects that are not in the 133 categories. However, OPSNet could extract their masks and give reasonable category proposals, like `mantle, gown, robe' for the `clothes'.
In row~2, we test on Berkeley dataset~\cite{berkeley}, OPSNet successfully predicts the `penguin' and the `leopard', which are not included in COCO. However, the prediction inevitably contains some noise. For example, in case (2) of Fig.~\ref{fig:demo1}, our model predicts the background as `rock' and `stone', but the `iceberg' is still within the top-5 predictions. 

\vspace{-4mm}
\paragraph{Hierarchical category prediction.} \label{hie}
WordNet~\cite{wordnet} gives the hierarchy for large amounts of vocabulary, which provides a better way to understand the world. Inspired by this, we explore building a hierarchical concept set.

In Fig.~\ref{fig:hie}, we make predictions with hierarchy via building a category tree.  
For example, when dealing with `Target 1' in the first row. We first make classification among coarse-grained categories like `thing' and `stuff', and gradually dive into some fine-grained categories like the specific types of cats. Finally, for this cat, we predict different levels of category [thing, animal, cat, siamese cat].

\vspace{-3pt}
\section{Conclusion}  
\vspace{-3pt}
We investigate open-vocabulary panoptic segmentation and propose a powerful solution named OPSNet. We develop exquisite designs like Embedding Modulation, Spatial Adapter and Mask Pooling, Mask Filtering, and Decoupled Supervision. The superior quantitative and qualitative results demonstrate its effectiveness and generality.

{\small
\bibliographystyle{ieee_fullname}
\bibliography{egbib}
}


\end{document}